\documentclass[10pt, sigconf]{acmart}

\AtBeginDocument{%
  \providecommand\BibTeX{{%
    \normalfont B\kern-0.5em{\scshape i\kern-0.25em b}\kern-0.8em\TeX}}}

\usepackage[ruled,vlined]{algorithm2e}
\usepackage[noend]{algpseudocode}

\usepackage{xspace}

\usepackage{tabularx}
\usepackage{amsthm}

\usepackage{amsmath}
\usepackage{dsfont}
\usepackage{wrapfig}
\usepackage[table]{colortbl}
\usepackage{makecell}
\usepackage{caption}

\usepackage{graphicx}
\usepackage{textcomp}
\usepackage{colortbl}
\usepackage{standalone}
\usepackage{booktabs}
\usepackage[shortlabels]{enumitem}

\definecolor{LightCyan}{rgb}{0.88,1,1}

\usepackage{multirow}
\usepackage{pgfplots}

\usepackage[nolist,nohyperlinks]{acronym}

\usepackage{subcaption}

\newcommand{\sysname}{\textbf{PluralLLM}\xspace}

\setcopyright{none}
\copyrightyear{2025}
\acmYear{2025}
\acmDOI{XXXXXXX.XXXXXXX}
\acmConference[ACM'25]{ACM Conference}{2025}{ }

\makeatletter
\renewcommand\@formatdoi[1]{\ignorespaces}
\makeatother

\begin{document}

\title[PluralLLM]{PluralLLM: Pluralistic Alignment in LLMs via Federated Learning}


\author{Mahmoud Srewa}
\email{msrewa@uci.edu}
\affiliation{%
  \institution{University of California, Irvine}
  \city{}
  \state{}
  \country{}
}
\author{Tianyu Zhao}
\email{tzhao15@uci.edu}
\affiliation{%
  \institution{University of California, Irvine}
  \city{}
  \state{}
  \country{}
}

\author{Salma Elmalaki}
\email{salma.elmalaki@uci.edu}
\affiliation{%
  \institution{University of California, Irvine}
  \city{}
  \state{}
  \country{}
}

\renewcommand{\shortauthors}{Srewa et al.}

\begin{acronym}
\acro{llms}[LLMs]{Large Language Models}
    \acro{llm}[LLM]{Large Language Model}
    \acro{fl}[FL]{Federated Learning}
    \acro{gpo}[GPO]{Group Preference Optimization}
    \acro{rlhf}[RLHF]{Reinforcement Learning from Human Feedback}
    \acro{dpo}[DPO]{Direct Preference Optimization}
    
\end{acronym}

\begin{abstract}
Ensuring \ac{llms} align with diverse human preferences while preserving privacy and fairness remains a challenge. Existing methods, such as \ac{rlhf}, rely on centralized data collection, making them computationally expensive and privacy-invasive. We introduce \sysname\footnote{The code will be released.} a federated learning-based approach that enables multiple user groups to collaboratively train a transformer-based preference predictor without sharing sensitive data, which can also serve as a reward model for aligning LLMs. Our method leverages Federated Averaging (FedAvg) to aggregate preference updates efficiently, achieving $46\%$ faster convergence, a $4\%$ improvement in alignment scores, and nearly the same group fairness measure as in centralized training. Evaluated on a Q/A preference alignment task, \sysname demonstrates that federated preference learning offers a scalable and privacy-preserving alternative for aligning LLMs with diverse human values.
\end{abstract}






\maketitle

\section{Introduction}\label{intro}


\ac{llms} have rapidly emerged as a cornerstone of modern artificial intelligence, powering applications ranging from conversational agents to content generation and decision support systems \cite{intro-survey2}. Their ability to generate human-like text has revolutionized various industries, but their effectiveness depends on their ability to align with human values and societal expectations~\cite{finetuning}. However, achieving robust human alignment remains a significant challenge, particularly in ensuring that these models fairly represent diverse perspectives, a concept known as Pluralistic Alignment \cite{pluralistic}.


Existing LLM alignment methods fall into two categories: prompt engineering and gradient-based alignment 
\cite{prompt,finetuning}. 
While prompt engineering guides model behavior through carefully crafted prompts and in-context examples without modifying model parameters, it often struggles with complex behaviors~\cite{prompot2}. Gradient-based alignment fine-tunes models using reward mechanisms, such as Reinforcement Learning from Human Feedback (\ac{rlhf}), improving traits like honesty and helpfulness but requiring extensive supervision and high computational costs~\cite{rlhf}. However, existing approaches do not scale efficiently for aligning LLMs across multiple user groups with limited supervision, making pluralistic alignment challenging~\cite{maxminrlf,rlhfissues}.



Despite recent advancements in alignment techniques, preference learning for \ac{llm} faces three significant challenges: (1) privacy risks, (2) collection of preference data, and (3) computational overhead~\cite{fl,rlhfissues}. Centralized approaches, such as \ac{rlhf} and gradient-based alignment, require collecting and processing user interactions, raising concerns about data security and user confidentiality. On the other hand, \ac{fl} enables privacy-preserving model training by keeping data decentralized. However, it comes with high communication and processing costs due to frequent updates between clients and global server~\cite{fl} . These challenges are even greater for pre-trained \ac{llm}, which require significant computational resources to incorporate preference data. Aligning \ac{llm} with FL adds further complexity, demanding a balance between alignment quality, efficiency, and privacy. In preference learning, this burden is particularly heavy, making efficient aggregation strategies or alternative models essential \cite{ fedbiot}. In this paper, we address the following research question:
\vspace{0.1cm}

\textbf{\textit{How can we align LLMs to capture the preference of various perspectives of different communities while preserving privacy, maintaining fairness, and ensuring computational efficiency?}}
\vspace{0.1cm}


We introduce \sysname, a framework for pluralistic alignment in LLMs via FL. Our approach leverages \ac{fl} to train a transformer-based preference predictor~\cite{gpo} to capture group-specific preferences in a distributed privacy-preserving manner. This preference predictor serves as a lightweight alternative to conventional alignment methods that require training a reward model \ac{rlhf}, significantly reducing computational overhead and can adapt to a new unseen group. Unlike the centralized group preference optimization training approach proposed by Zhao et. al.~\cite{gpo}, our FL-based preference predictor ensures privacy-preserving preference learning by allowing different groups to collaboratively train the model without exposing their sensitive preference data. In addition, it enables diverse groups to participate in training while maintaining the fairness properties of the centralized approach. Our results demonstrate that our proposed \sysname approach  
applied to preference learning achieves higher alignment scores and faster convergence compared to centralized methods, making it a scalable and efficient solution for capturing diverse group preferences.
\vspace{-0.3cm}

\begin{figure*}[!t]
\centering
\includegraphics[width=\textwidth]{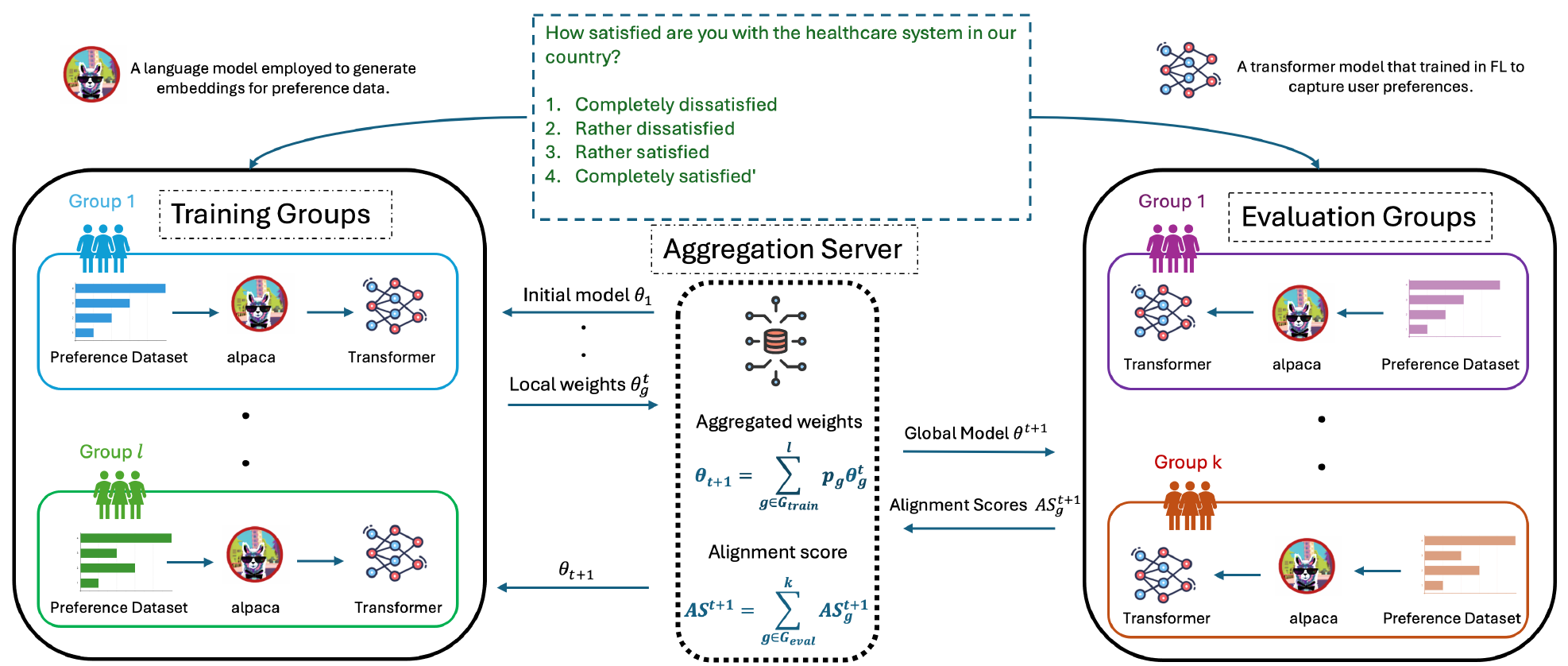}
 \caption{\sysname: Pluralistic alignment in LLMs via Federated Learning.}
\label{fig:framework}
\end{figure*}
\section{Related Work}\label{sec:related}
\hspace{\parindent} \textbf{Prompt Engineering}: Prompt engineering provides a mechanism for fine-tuning model outputs through the modification of the input to the \ac{llm}, thereby aligning with user preferences without altering the parameters of the core \ac{llm} model. Prompt engineering approaches are characterized by their computational efficiency, a property that stems from the absence of any training requirements~\cite{prompt}. However, the design of the prompt itself can be a laborious task that relies on heuristics. The efficacy of these heuristics is not guaranteed to transfer well across different LLMs~\cite{gpo}. Recent work in the literature has shown that prompt engineering has limited success in aligning LLMs to complex groups on challenging survey datasets such as ~\textit{GlobalOpinionQA}~\cite{durmus2023towards}.

\textbf{Pluralistic Alignment in LLM:}
A growing number of pluralistic alignment studies show that it is important to design \ac{llm} systems that can accommodate and represent diverse human values, perspectives, and preferences. Unlike traditional alignment approaches that aim to align models to a single, averaged set of human preferences, pluralistic alignment seeks to reflect the complexity and plurality of human societies. For example, Cao et. al. introduced an age fairness reward in \ac{llm} to reduce response quality disparities across distinct age groups during training~\cite{cao2024agr}. Traditional Reinforcement Learning alignment approaches, such as \ac{rlhf} often reinforce majority viewpoints while marginalizing minority perspectives. The question of balancing openness to diverse values with ethical constraints, such as the avoidance of harmful ideologies, remains largely unaddressed.

\textbf{Group Preference Alignment: }
Group preference alignment refers to techniques designed to adapt \ac{llm} outputs to reflect the distinct preferences, values, or judgments of different groups or demographics. \ac{gpo}~\cite{gpo} was introduced as a few-shot alignment framework that steers LLMs toward group-specific preferences. \ac{gpo} augments the base LLM with an independent transformer module, trained via in-context supervised learning with only a handful of samples to predict group preferences and refine model outputs. This module acts as a preference model for different groups, learning distinct alignment patterns across diverse communities. By leveraging an in-context autoregressive transformer, \ac{gpo} enables flexible and efficient alignment, allowing LLMs to adapt dynamically to varying user preferences.
\vspace{-0.2cm}

\section{Pluralistic Alignment in LLMs via Federated Learning}\label{sec:threatmodel}

We chose the Q/A preference alignment task, which involves aligning \ac{llm} responses based on group-specific preferences. This task is particularly well-suited for evaluating pluralistic alignment, as it requires the model to adapt to diverse user opinions while maintaining coherence and fairness. Unlike standard classification tasks, preference-based Q/A alignment provides a richer evaluation metric, allowing us to measure not only the correctness of responses but also how well the model captures nuanced group preferences. This setup also reflects real-world applications, where \ac{llm} must personalize responses based on collective user preferences. We adopt a \ac{fl} setup for pluralistic alignment in \ac{llm}. The framework consists of three main actors, as described in \autoref{fig:framework}:

\begin{itemize}[topsep=0pt, leftmargin=*]
  \item \textbf{Training Clients (Groups):} The training set, $G_{\text{train}}$, comprises $l$ distinct groups, each representing a client in a FL setup. Each client performs local training to develop a transformer-based preference model \cite{gpo}. This model aims to learn group-specific preference patterns and generalize to unseen data. Each client trains its transformer independently using its respective group's preference dataset. 

    \item \textbf{Aggregation Server:} A central server coordinates \ac{fl} by collecting, aggregating, and redistributing model updates from training clients. This process enables learning from diverse data while preserving privacy and preventing direct data sharing. 

  \item \textbf{Evaluation Clients (Groups):} A separate set $G_{\text{eval}}$ consisting of $K$ groups is introduced, where each group acts as a client in the FL setup to assess the alignment performance of the trained model. Unlike training clients, these groups do not participate in model updates. Instead, they represent new unseen groups that serve as an independent benchmark to evaluate the generalizability of the trained model. Their feedback helps determine how well the aggregated model aligns with unseen groups.

\end{itemize}

\vspace{-0.2cm}

\subsection{Group Local Training}
Each training group $g \in G_{\text{train}}$ has a preference dataset: $D_g = \{(x_1^g, y_1^g), \dots, (x_n^g, y_n^g)\}$ where $x_i^g$ represents the embedding of \ac{llm}  concatenated prompt-response pair: $x_i^g = \omega_{\text{emb}}(q_i^g, r_i^g)$ and $\omega_{\text{emb}}$ is a language model embedding function. The preference $y_i^g$ represents the group preference probability for the generated response $r_i^g$ to the query $q_i^g$. For instance, in \autoref{fig:framework}, the query represents the question, while the response corresponds to the aggregated probability distribution of answers within a group. This preference data is then processed by the Alpaca model (\ac{llm}) to generate embeddings, following the approach in \cite{gpo}.

The dataset for each group is randomly divided into $m$ context samples $(x_1^g, y_1^g), ..., (x_m^g, y_m^g)$ and $n-m$ target samples $(x_{m+1}^g, y_{m+1}^g),.., (x_n^g, y_n^g)$. The model is trained to predict the target preferences given the context examples, optimizing the following loss function as introduced in \cite{gpo}:

\begin{equation}
\mathcal{L}(\theta) = \mathbb{E}_{g,m} \left[ \sum_{i=m+1}^{n} \log p_{\theta} (y_i^g \mid x_{1:m}^g, y_{1:m}^g, x_i^g) \right],
\label{eq:localloss}
\end{equation}

where $p_{\theta}$ denotes the target points predicted preference distribution conditioned on the context points. At the end of local training, each client $g$ transmits its updated model parameters $\theta_g$ to the central aggregation server, which combines the received updates to train the global model.








\subsection{Model Aggregation}
We employ the FedAvg technique to aggregate updates from multiple groups~\cite{fedavg}. The aggregation process is designed to minimize the global optimization function of FL. 

\begin{equation}
\min_{\theta}\ F(\theta) = \sum_{g \in G_{\text{train}}} p_g F_g(\theta),
\end{equation}

where \( p_g \) is the weight assigned to each training group \( g \), defined as: $p^g = \frac{|D_g|}{\sum_{g'} |D_{g'}|}$. Here, \( D_g \) represents the size of the preference dataset for group \( g \). The term \( p_g \) ensures that each group's contribution to the global objective is weighted by the proportion of its dataset size relative to the total dataset across all training groups. The local objective function for group \( g \) is defined as: $
F_g(\theta) = \mathbb{E}_{(x, y) \sim D_g} \left[ \mathcal{L} (f_{\theta}(x), y) \right]$, where $\mathcal{L}$ represents the loss function defined at \autoref{eq:localloss} and $f_{\theta}(x)$ is the model output for input $x$. To approximate the global objective, model updates from training groups are aggregated as follows:

\begin{equation}
\theta^{t+1} = \sum_{g \in G_{\text{train}}} p_g \theta_g^t,
\end{equation}

where $\theta_g^t$ represents the locally updated model parameters at the training group $g$ in round $t$. The aggregated model is then redistributed to training and evaluation clients.





\section{Experiments}\label{sec:eval} 


\begin{figure}[!t]
\centering
\includegraphics[width=0.9\linewidth]{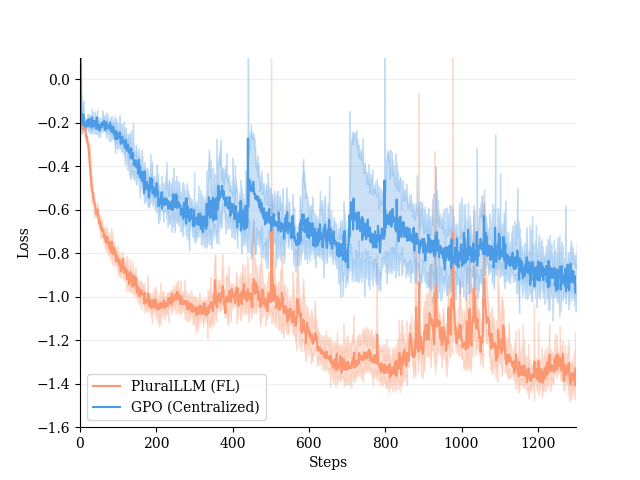}
\caption{Comparison of training loss curves for centralized learning GPO and \sysname. \sysname achieves a lower loss compared to Centralized Training GPO.}
\label{fig:loss}
\end{figure}

\begin{figure}[!t]
\centering
\includegraphics[width=1\linewidth]{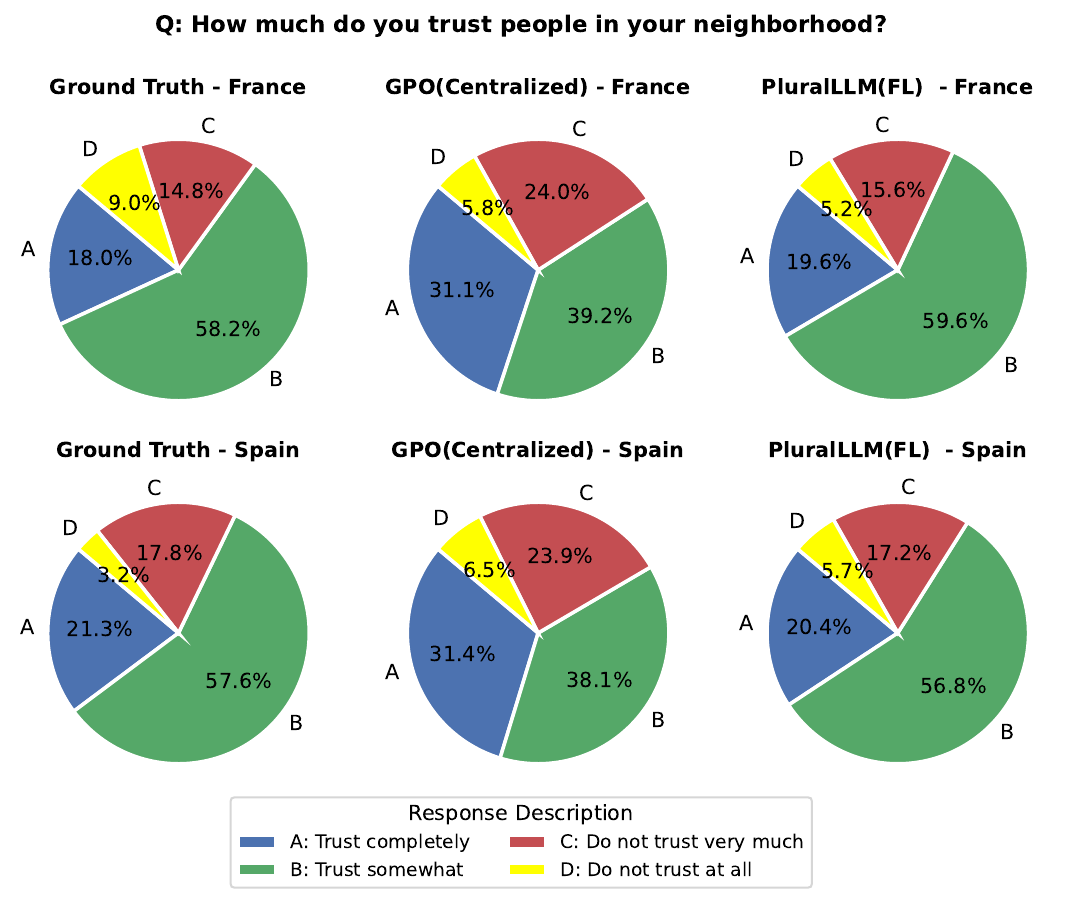}
\caption{Comparison of preference distributions across Ground Truth, Centralized Learning GPO, and \sysname for a given question.}
\label{fig:pie}
\end{figure}

\subsection{Experimental Setup} 
Our experiments utilize a transformer-based preference predictor model (\ac{gpo} ~\cite{gpo}), originally designed to train groups sequentially in an ordered manner. However, in \sysname, we adapt it to a FL setting, employing FedAvg as a completely different learning paradigm. The primary goal of our evaluation is to determine whether our approach in \sysname impacts the alignment score and group fairness across different groups compared to the original centralized learning. 

Experiments are conducted on machines equipped with one NVIDIA A30 GPU, an Intel(R) Xeon(R) Gold 6326 CPU @ 2.90GHz, and 256GB RAM. All results reported in this study are averaged over four different runs with varying random seeds to ensure stability.

\subsection{Dataset}
We used the dataset from the Pew Research Center's Global Attitudes Surveys (PewResearch), which collects public opinions on a wide range of social, political, and economic issues \cite{globalopnion}. These surveys capture diverse perspectives from various demographic and geographic groups, providing a rich foundation for analyzing preference alignment. To ensure a fair comparison with \ac{gpo}~\cite{gpo}, we use the same subset of groups as in the original \ac{gpo} training setup. In both \ac{fl} and centralized learning, groups are partitioned into $60\%$  and $40\%$ groups for training and evaluation, respectively.

\subsection{Implementation}
\begin{itemize}[topsep=0pt, leftmargin=*]
    \item 
\textbf{Federated Learning:} We trained the transformer-based preference predictor model (\ac{gpo}) for $1,300$ communication rounds in the \ac{fl} setup, assuming that all clients participate in each communication round. Each local training step consists of 6 local epochs, where, in each epoch, we randomly sample context questions and corresponding preferences, then the target questions that we wish to predict its preferences to train the model. Adam optimizer used with a learning rate of $3 \times 10^{-4}$.
    \item 
\textbf{Centralized Learning:} We trained the transformer-based preference predictor model (\ac{gpo}) for $1,300$ epochs, iterating over all training groups in each epoch. During the training epoch, each group samples a random selection of context questions and their corresponding preferences and target questions.  
Unlike \ac{fl}, where model updates are aggregated after each communication round, the centralized approach updates the model sequentially for each group within a single epoch. 
    \item 
\textbf{Preferences Embedding}: We use Alpaca-7B, a fine-tuned version of LLaMA-7B, as the embedding model to represent preference data, which is then passed to transformer input~\cite{gpo}. The embedding step is done once over all the preference data for each group at the beginning of training.
\end{itemize}
To assess alignment performance, evaluation is conducted every $10$ communication rounds (in the \ac{fl} setting) or every $10$ epochs (in the centralized setting). The alignment score is computed over randomly sampled data from all the evaluation groups to measure how well the trained model adapts to new group preferences over time.

\subsection{Evaluation Metrics}
To quantify the impact of \sysname on alignment performance, we evaluate:
\begin{itemize}[topsep=0pt, leftmargin=*]
    \item \textbf{Alignment Score(\texttt{AS}):} We assess the degree of alignment between $2$ opinion distributions $P_1$ and $P_2$ by calculating the Alignment Score \texttt{$AS(P_1, P_2; Q)$} over a set of questions $Q$ as used in ~\cite{gpo}. Jensen-Shannon Distance \footnote{The Jensen–Shannon divergence (JSD) is a symmetric measure of similarity between two probability distributions, always non-negative, with $0$ denoting identical distributions and any value above $0$ indicating differences.}, denoted as \textit{JSD}, is used to assess preference distribution similarity shifts.
    \begin{equation}
        AS(P_1, P_2; Q) = \frac{1}{|Q|}\sum_{q\in Q}\text{\textit{JSD}}(P_1(q), P_2(q); Q)
    \end{equation}
    
    \item \textbf{Convergence Speed of Loss Function: } We measure how quickly the model optimizes preference alignment by tracking the average loss across all clients at each communication round. This is compared to the centralized training loss per epoch. Convergence speed is defined as the point where the model reaches $95\%$ of its final loss value. 
    

    \item \textbf{Fairness Metrics:} We analyze the effect of \sysname on group fairness in preference alignment across different groups compared to centralized learning. Numerous studies have demonstrated that FL can inadvertently introduce unfairness into the trained models~\cite{shi2023towards}. 
    This unfairness arises primarily from data heterogeneity across clients, which leads to disparate performance results and challenges in achieving equitable model accuracy across all participants. Furthermore, these fairness issues have the potential to show disparity in privacy leakage risks, as adversaries can potentially exploit shared model parameters to infer sensitive information~\cite{zhao2025finp}. We assess the fairness by adapting Coefficient of Variation (CoV) and Fairness Index(FI) to measure the disparity of alignment scores across distinct groups. These metrics are used to measure the relative perception of fairness in human-centered systems~\cite{zhao2024fina, zhao2024fairo}. 
    For $K$ groups, we define the alignment score of group $i$ as $AS_i$. The average alignment score across groups is calculated as $\mu = \frac{1}{K} \sum_{i=1}^{K} \text{AS}_i$. The CoV of alignment scores $CoV(AS)$ is calculated as:
    \begin{align}\label{eq:covas}
    CoV(AS) = \frac{\sigma}{\mu} = \frac{\sqrt{\frac{1}{K} \sum_{i=1}^{K} (\text{AS}_i - \mu)^2}}{\mu},
    \end{align}
    where $\sigma$ is the standard derivation of $AS$. A lower CoV indicates a more equitable distribution of alignment scores among groups, suggesting better fairness in aligning distinct groups. We apply the Fairness Index (FI) transformation to interpret the fairness in percentage between $0$ and $1$. A higher FI indicates greater fairness, where $1$ represents perfect fairness.
    \begin{align}\label{eq:fias}
    FI(AS) = \frac{1}{1 + CoV^2(AS)} 
    \end{align}
    In the context of group fairness in Machine Learning (ML) classification task, the principle of equal opportunity implies that individuals who are eligible for a favorable outcome have an equal chance of being correctly classified by the prediction model, regardless of their group membership. Similar to the classification task, we investigate the equal opportunity in \ac{llm} alignment task. 
    To adapt the definition of equal opportunity in the probabilistic setting of LLM alignment, we use our definition of $CoV$ (\autoref{eq:covas}). In this context, equal opportunity would imply that the variability of alignment scores between groups is minimum or the $FI$ (\autoref{eq:fias}) is close to $1$.
\end{itemize}

\subsection{Analysis of Convergence Speed}
\sysname converges at communication round $634$, whereas the centralized approach requires significantly more steps, converging at iteration $1180$ epoch as seen in \autoref{fig:loss}. Hence, \sysname achieves convergence $\textbf{46\%}$ faster than the centralized learning approach, highlighting its efficiency in accelerating model training. Additionally, \sysname maintains a lower loss throughout training as observed in \autoref{fig:loss}, demonstrating improved stability compared to the centralized approach. The faster convergence of \sysname suggests that it is well-suited for distributed learning scenarios where reducing communication rounds is critical for efficiency.


\subsection{Analysis of Alignment Performance}

\begin{figure}[t]
    \centering
    \begin{subfigure}{0.45\textwidth}
        \centering
        \includegraphics[width=0.9\linewidth]{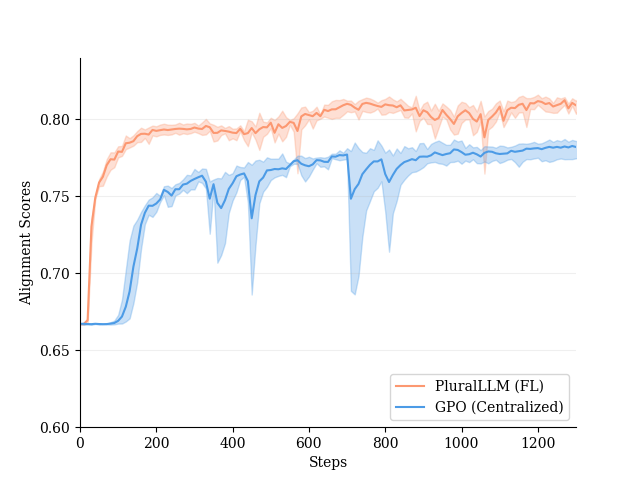}
    \end{subfigure}
    \caption{Comparison of mean evaluation group alignment scores for centralized learning GPO and \sysname. }
    \label{fig:align}
\end{figure}
 \autoref{fig:align} demonstrates that \sysname achieves a $\approx 4$\% improvement in the mean evaluation alignment score compared to the centralized approach. While the centralized method shows slower improvements and remains at a lower score, \sysname maintains a more stable progression with lower fluctuations, indicating that the distributed training approach enhances robustness in achieving better alignment and improves generalization across diverse data distributions. Additionally, \autoref{fig:pie} highlights that for two evaluation groups, \sysname more accurately represents the baseline distribution for a given Q/A task compared to the centralized approach.

\subsection{Analysis of Fairness in Alignment}

\begin{figure}[t]
    \centering
    \begin{subfigure}{0.45\textwidth}
        \centering
        \includegraphics[width=0.9\linewidth]{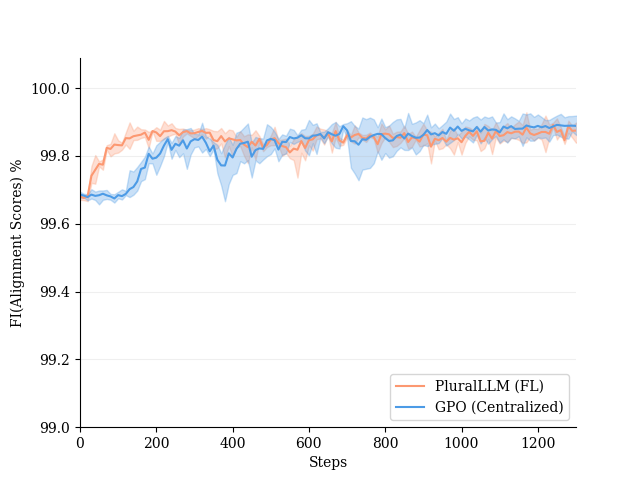}
    \end{subfigure}
    \caption{Comparison of Fairness Index between centralized learning and \sysname. The utilization of \sysname in the training of a preference transformer \textit{Does Not} result in significant disparities among groups.}
    \label{fig:FI}
\end{figure}

As observed in ~\autoref{fig:FI}, \sysname improves the $FI$ by $0.04\%$ on average before converging at a round $634$ compared to the centralized GPO. \sysname maintains a comparable $FI$ across training steps till round $1300$ achieving an equal opportunity with $FI\approx1$. 


\section{Discussion \& Conclusion}

\sysname introduces a federated learning-based approach for pluralistic alignment in LLMs, addressing privacy, efficiency, and scalability challenges. Decentralizing the training of a transformer-based preference predictor preserves user privacy while capturing diverse group preferences more effectively than centralized methods. Our evaluation employs FedAvg for efficient preference update aggregation, resulting in $46\%$ faster convergence, a $4$\% improvement in alignment scores, and maintaining group fairness comparable to centralized training. 

In addition, this predictor can serve as a lightweight reward function for RLHF, reducing computational costs or generating high-quality preference datasets for DPO, improving efficiency. While effective in Q/A tasks, its applicability to other domains like summarization and translation remains unexplored. Future work will focus on integrating learned preferences into LLM fine-tuning methods and exploring alternative aggregation strategies to enhance fairness across tasks. Furthermore, extending \sysname beyond Q/A to diverse learning tasks.


\begin{acks}
This work is supported by the U.S. National Science Foundation (NSF) under grant number 2339266.
\end{acks}

\balance
\bibliographystyle{ACM-Reference-Format}
\bibliography{humansysllm}


\begin{thebibliography}{19}


\ifx \showCODEN    \undefined \def \showCODEN     #1{\unskip}     \fi
\ifx \showISBNx    \undefined \def \showISBNx     #1{\unskip}     \fi
\ifx \showISBNxiii \undefined \def \showISBNxiii  #1{\unskip}     \fi
\ifx \showISSN     \undefined \def \showISSN      #1{\unskip}     \fi
\ifx \showLCCN     \undefined \def \showLCCN      #1{\unskip}     \fi
\ifx \shownote     \undefined \def \shownote      #1{#1}          \fi
\ifx \showarticletitle \undefined \def \showarticletitle #1{#1}   \fi
\ifx \showURL      \undefined \def \showURL       {\relax}        \fi
\providecommand\bibfield[2]{#2}
\providecommand\bibinfo[2]{#2}
\providecommand\natexlab[1]{#1}
\providecommand\showeprint[2][]{arXiv:#2}

\bibitem[Bubeck et~al\mbox{.}(2023)]%
        {intro-survey2}
\bibfield{author}{\bibinfo{person}{S{\'e}bastien Bubeck}, \bibinfo{person}{Varun Chandrasekaran}, \bibinfo{person}{Ronen Eldan}, \bibinfo{person}{Johannes Gehrke}, \bibinfo{person}{Eric Horvitz}, \bibinfo{person}{Ece Kamar}, \bibinfo{person}{Peter Lee}, \bibinfo{person}{Yin~Tat Lee}, \bibinfo{person}{Yuanzhi Li}, \bibinfo{person}{Scott Lundberg}, {et~al\mbox{.}}} \bibinfo{year}{2023}\natexlab{}.
\newblock \showarticletitle{Sparks of artificial general intelligence: Early experiments with gpt-4}.
\newblock \bibinfo{journal}{\emph{arXiv preprint arXiv:2303.12712}} (\bibinfo{year}{2023}).
\newblock


\bibitem[Cao et~al\mbox{.}(2024)]%
        {cao2024agr}
\bibfield{author}{\bibinfo{person}{Shuirong Cao}, \bibinfo{person}{Ruoxi Cheng}, {and} \bibinfo{person}{Zhiqiang Wang}.} \bibinfo{year}{2024}\natexlab{}.
\newblock \showarticletitle{AGR: Age Group fairness Reward for Bias Mitigation in LLMs}.
\newblock \bibinfo{journal}{\emph{arXiv preprint arXiv:2409.04340}} (\bibinfo{year}{2024}).
\newblock


\bibitem[Casper et~al\mbox{.}(2023)]%
        {rlhfissues}
\bibfield{author}{\bibinfo{person}{Stephen Casper}, \bibinfo{person}{Xander Davies}, \bibinfo{person}{Claudia Shi}, \bibinfo{person}{Thomas~Krendl Gilbert}, \bibinfo{person}{J{\'e}r{\'e}my Scheurer}, \bibinfo{person}{Javier Rando}, \bibinfo{person}{Rachel Freedman}, \bibinfo{person}{Tomasz Korbak}, \bibinfo{person}{David Lindner}, \bibinfo{person}{Pedro Freire}, {et~al\mbox{.}}} \bibinfo{year}{2023}\natexlab{}.
\newblock \showarticletitle{Open problems and fundamental limitations of reinforcement learning from human feedback}.
\newblock \bibinfo{journal}{\emph{arXiv preprint arXiv:2307.15217}} (\bibinfo{year}{2023}).
\newblock


\bibitem[Chakraborty et~al\mbox{.}(2024)]%
        {maxminrlf}
\bibfield{author}{\bibinfo{person}{Souradip Chakraborty}, \bibinfo{person}{Jiahao Qiu}, \bibinfo{person}{Hui Yuan}, \bibinfo{person}{Alec Koppel}, \bibinfo{person}{Furong Huang}, \bibinfo{person}{Dinesh Manocha}, \bibinfo{person}{Amrit~Singh Bedi}, {and} \bibinfo{person}{Mengdi Wang}.} \bibinfo{year}{2024}\natexlab{}.
\newblock \showarticletitle{MaxMin-RLHF: Alignment with diverse human preferences}.
\newblock \bibinfo{journal}{\emph{arXiv preprint arXiv:2402.08925}} (\bibinfo{year}{2024}).
\newblock


\bibitem[Durmus et~al\mbox{.}(2023a)]%
        {durmus2023towards}
\bibfield{author}{\bibinfo{person}{Esin Durmus}, \bibinfo{person}{Karina Nguyen}, \bibinfo{person}{Thomas~I Liao}, \bibinfo{person}{Nicholas Schiefer}, \bibinfo{person}{Amanda Askell}, \bibinfo{person}{Anton Bakhtin}, \bibinfo{person}{Carol Chen}, \bibinfo{person}{Zac Hatfield-Dodds}, \bibinfo{person}{Danny Hernandez}, \bibinfo{person}{Nicholas Joseph}, {et~al\mbox{.}}} \bibinfo{year}{2023}\natexlab{a}.
\newblock \showarticletitle{Towards measuring the representation of subjective global opinions in language models}.
\newblock \bibinfo{journal}{\emph{arXiv preprint arXiv:2306.16388}} (\bibinfo{year}{2023}).
\newblock


\bibitem[Durmus et~al\mbox{.}(2023b)]%
        {globalopnion}
\bibfield{author}{\bibinfo{person}{Esin Durmus}, \bibinfo{person}{Karina Nguyen}, \bibinfo{person}{Thomas~I Liao}, \bibinfo{person}{Nicholas Schiefer}, \bibinfo{person}{Amanda Askell}, \bibinfo{person}{Anton Bakhtin}, \bibinfo{person}{Carol Chen}, \bibinfo{person}{Zac Hatfield-Dodds}, \bibinfo{person}{Danny Hernandez}, \bibinfo{person}{Nicholas Joseph}, {et~al\mbox{.}}} \bibinfo{year}{2023}\natexlab{b}.
\newblock \showarticletitle{Towards measuring the representation of subjective global opinions in language models}.
\newblock \bibinfo{journal}{\emph{arXiv preprint arXiv:2306.16388}} (\bibinfo{year}{2023}).
\newblock


\bibitem[Kuang et~al\mbox{.}(2024)]%
        {fl}
\bibfield{author}{\bibinfo{person}{Weirui Kuang}, \bibinfo{person}{Bingchen Qian}, \bibinfo{person}{Zitao Li}, \bibinfo{person}{Daoyuan Chen}, \bibinfo{person}{Dawei Gao}, \bibinfo{person}{Xuchen Pan}, \bibinfo{person}{Yuexiang Xie}, \bibinfo{person}{Yaliang Li}, \bibinfo{person}{Bolin Ding}, {and} \bibinfo{person}{Jingren Zhou}.} \bibinfo{year}{2024}\natexlab{}.
\newblock \showarticletitle{Federatedscope-llm: A comprehensive package for fine-tuning large language models in federated learning}. In \bibinfo{booktitle}{\emph{Proceedings of the 30th ACM SIGKDD Conference on Knowledge Discovery and Data Mining}}. \bibinfo{pages}{5260--5271}.
\newblock


\bibitem[McMahan et~al\mbox{.}(2017)]%
        {fedavg}
\bibfield{author}{\bibinfo{person}{Brendan McMahan}, \bibinfo{person}{Eider Moore}, \bibinfo{person}{Daniel Ramage}, \bibinfo{person}{Seth Hampson}, {and} \bibinfo{person}{Blaise~Aguera y Arcas}.} \bibinfo{year}{2017}\natexlab{}.
\newblock \showarticletitle{Communication-efficient learning of deep networks from decentralized data}. In \bibinfo{booktitle}{\emph{Artificial intelligence and statistics}}. PMLR, \bibinfo{pages}{1273--1282}.
\newblock


\bibitem[Ouyang et~al\mbox{.}(2022)]%
        {rlhf}
\bibfield{author}{\bibinfo{person}{Long Ouyang}, \bibinfo{person}{Jeffrey Wu}, \bibinfo{person}{Xu Jiang}, \bibinfo{person}{Diogo Almeida}, \bibinfo{person}{Carroll Wainwright}, \bibinfo{person}{Pamela Mishkin}, \bibinfo{person}{Chong Zhang}, \bibinfo{person}{Sandhini Agarwal}, \bibinfo{person}{Katarina Slama}, \bibinfo{person}{Alex Ray}, {et~al\mbox{.}}} \bibinfo{year}{2022}\natexlab{}.
\newblock \showarticletitle{Training language models to follow instructions with human feedback}.
\newblock \bibinfo{journal}{\emph{Advances in neural information processing systems}}  \bibinfo{volume}{35} (\bibinfo{year}{2022}), \bibinfo{pages}{27730--27744}.
\newblock


\bibitem[Sahoo et~al\mbox{.}(2024)]%
        {prompt}
\bibfield{author}{\bibinfo{person}{Pranab Sahoo}, \bibinfo{person}{Ayush~Kumar Singh}, \bibinfo{person}{Sriparna Saha}, \bibinfo{person}{Vinija Jain}, \bibinfo{person}{Samrat Mondal}, {and} \bibinfo{person}{Aman Chadha}.} \bibinfo{year}{2024}\natexlab{}.
\newblock \showarticletitle{A systematic survey of prompt engineering in large language models: Techniques and applications}.
\newblock \bibinfo{journal}{\emph{arXiv preprint arXiv:2402.07927}} (\bibinfo{year}{2024}).
\newblock


\bibitem[Shi et~al\mbox{.}(2023)]%
        {shi2023towards}
\bibfield{author}{\bibinfo{person}{Yuxin Shi}, \bibinfo{person}{Han Yu}, {and} \bibinfo{person}{Cyril Leung}.} \bibinfo{year}{2023}\natexlab{}.
\newblock \showarticletitle{Towards fairness-aware federated learning}.
\newblock \bibinfo{journal}{\emph{IEEE Transactions on Neural Networks and Learning Systems}} (\bibinfo{year}{2023}).
\newblock


\bibitem[Sorensen et~al\mbox{.}(2024)]%
        {pluralistic}
\bibfield{author}{\bibinfo{person}{Taylor Sorensen}, \bibinfo{person}{Jared Moore}, \bibinfo{person}{Jillian Fisher}, \bibinfo{person}{Mitchell Gordon}, \bibinfo{person}{Niloofar Mireshghallah}, \bibinfo{person}{Christopher~Michael Rytting}, \bibinfo{person}{Andre Ye}, \bibinfo{person}{Liwei Jiang}, \bibinfo{person}{Ximing Lu}, \bibinfo{person}{Nouha Dziri}, {et~al\mbox{.}}} \bibinfo{year}{2024}\natexlab{}.
\newblock \showarticletitle{A roadmap to pluralistic alignment}.
\newblock \bibinfo{journal}{\emph{arXiv preprint arXiv:2402.05070}} (\bibinfo{year}{2024}).
\newblock


\bibitem[Wu et~al\mbox{.}(2024)]%
        {fedbiot}
\bibfield{author}{\bibinfo{person}{Feijie Wu}, \bibinfo{person}{Zitao Li}, \bibinfo{person}{Yaliang Li}, \bibinfo{person}{Bolin Ding}, {and} \bibinfo{person}{Jing Gao}.} \bibinfo{year}{2024}\natexlab{}.
\newblock \showarticletitle{Fedbiot: Llm local fine-tuning in federated learning without full model}. In \bibinfo{booktitle}{\emph{Proceedings of the 30th ACM SIGKDD Conference on Knowledge Discovery and Data Mining}}. \bibinfo{pages}{3345--3355}.
\newblock


\bibitem[Zhang et~al\mbox{.}(2023)]%
        {finetuning}
\bibfield{author}{\bibinfo{person}{Shengyu Zhang}, \bibinfo{person}{Linfeng Dong}, \bibinfo{person}{Xiaoya Li}, \bibinfo{person}{Sen Zhang}, \bibinfo{person}{Xiaofei Sun}, \bibinfo{person}{Shuhe Wang}, \bibinfo{person}{Jiwei Li}, \bibinfo{person}{Runyi Hu}, \bibinfo{person}{Tianwei Zhang}, \bibinfo{person}{Fei Wu}, {et~al\mbox{.}}} \bibinfo{year}{2023}\natexlab{}.
\newblock \showarticletitle{Instruction tuning for large language models: A survey}.
\newblock \bibinfo{journal}{\emph{arXiv preprint arXiv:2308.10792}} (\bibinfo{year}{2023}).
\newblock


\bibitem[Zhao et~al\mbox{.}(2023)]%
        {gpo}
\bibfield{author}{\bibinfo{person}{Siyan Zhao}, \bibinfo{person}{John Dang}, {and} \bibinfo{person}{Aditya Grover}.} \bibinfo{year}{2023}\natexlab{}.
\newblock \showarticletitle{Group preference optimization: Few-shot alignment of large language models}.
\newblock \bibinfo{journal}{\emph{arXiv preprint arXiv:2310.11523}} (\bibinfo{year}{2023}).
\newblock


\bibitem[Zhao and Elmalaki(2024)]%
        {zhao2024fina}
\bibfield{author}{\bibinfo{person}{Tianyu Zhao} {and} \bibinfo{person}{Salma Elmalaki}.} \bibinfo{year}{2024}\natexlab{}.
\newblock \showarticletitle{FinA: Fairness of adverse effects in decision-making of human-cyber-physical-system}. In \bibinfo{booktitle}{\emph{2024 ACM/IEEE 15th International Conference on Cyber-Physical Systems (ICCPS)}}. IEEE, \bibinfo{pages}{202--211}.
\newblock


\bibitem[Zhao et~al\mbox{.}(2025)]%
        {zhao2025finp}
\bibfield{author}{\bibinfo{person}{Tianyu Zhao}, \bibinfo{person}{Mahmoud Srewa}, {and} \bibinfo{person}{Salma Elmalaki}.} \bibinfo{year}{2025}\natexlab{}.
\newblock \showarticletitle{FinP: Fairness-in-Privacy in Federated Learning by Addressing Disparities in Privacy Risk}.
\newblock \bibinfo{journal}{\emph{arXiv preprint arXiv:2502.17748}} (\bibinfo{year}{2025}).
\newblock


\bibitem[Zhao et~al\mbox{.}(2024)]%
        {zhao2024fairo}
\bibfield{author}{\bibinfo{person}{Tianyu Zhao}, \bibinfo{person}{Mojtaba Taherisadr}, {and} \bibinfo{person}{Salma Elmalaki}.} \bibinfo{year}{2024}\natexlab{}.
\newblock \showarticletitle{Fairo: Fairness-aware sequential decision making for human-in-the-loop cps}. In \bibinfo{booktitle}{\emph{2024 ACM/IEEE 15th International Conference on Cyber-Physical Systems (ICCPS)}}. IEEE, \bibinfo{pages}{87--98}.
\newblock


\bibitem[Zhou et~al\mbox{.}(2022)]%
        {prompot2}
\bibfield{author}{\bibinfo{person}{Yongchao Zhou}, \bibinfo{person}{Andrei~Ioan Muresanu}, \bibinfo{person}{Ziwen Han}, \bibinfo{person}{Keiran Paster}, \bibinfo{person}{Silviu Pitis}, \bibinfo{person}{Harris Chan}, {and} \bibinfo{person}{Jimmy Ba}.} \bibinfo{year}{2022}\natexlab{}.
\newblock \showarticletitle{Large language models are human-level prompt engineers}. In \bibinfo{booktitle}{\emph{The Eleventh International Conference on Learning Representations}}.
\newblock


\end{thebibliography}

\appendix

\end{document}